\pdfoutput=1

\documentclass[11pt,a4paper]{article}
\usepackage{times,latexsym}
\usepackage{url}
\usepackage[T1]{fontenc}

\usepackage[acceptedWithA]{tacl2018v2}

\usepackage{multirow}
\usepackage{booktabs}
\usepackage{soul}
\usepackage{color}
\usepackage{graphicx}
\usepackage{amsmath}
\usepackage{amssymb}
\usepackage{dsfont}
\usepackage{xcolor}
\usepackage{lipsum}
\usepackage{amsthm}
\usepackage{appendix}
\usepackage{dsfont}
\usepackage[autostyle]{csquotes}  
\usepackage{enumitem}

\DeclareMathOperator*{\argmin}{argmin}
\DeclareMathOperator*{\argmax}{argmax}
\usepackage[most]{tcolorbox}

\newcommand\cameraready[1]{\textcolor{purple}{}}
\newcommand\add[1]{#1}
\newcommand\remove[1]{}

\usepackage{xspace,mfirstuc,tabulary}

\newif\iftaclinstructions
\taclinstructionsfalse 
\iftaclinstructions
\renewcommand{\confidential}{}
\renewcommand{\anonsubtext}{(No author info supplied here, for consistency with
TACL-submission anonymization requirements)}
\newcommand{\instr}
\fi

\iftaclpubformat 

\else

\fi

\title{Aligning Faithful Interpretations with their Social Attribution}

\author{Alon Jacovi \\
  Bar Ilan University \\
  \texttt{alonjacovi@gmail.com} \\\And
  Yoav Goldberg \\
  Bar Ilan University and 
  Allen Institute for AI \\
  \texttt{yoav.goldberg@gmail.com}
  }

\date{}

\begin{document}
\maketitle
\begin{abstract}


We find that the requirement of model interpretations to be \emph{faithful} is vague and incomplete.
With interpretation by textual highlights as a case-study, we present several failure cases. Borrowing concepts from social science, we identify that the problem is a misalignment between the causal chain of decisions (causal attribution) and the attribution of human behavior to the interpretation (social attribution). We re-formulate faithfulness as an accurate attribution of causality to the model, and introduce the concept of \textit{aligned} \textit{faithfulness}: faithful causal chains that are aligned with their expected social behavior. The two steps of causal attribution and social attribution \textit{together} complete the process of explaining behavior.
With this formalization, we characterize various failures of misaligned faithful highlight interpretations, and propose an alternative causal chain to remedy the issues. Finally, we implement highlight explanations of the proposed causal format using contrastive explanations.

\end{abstract}

\section{Introduction}





When formalizing the desired properties of a quality interpretation of a model or a decision, the NLP community has settled on the key property of \textit{faithfulness} \cite{lipton2016mythos,DBLP:journals/corr/abs-1711-07414,wiegreffe2019attentionisnotnot,jacovi2020}, or how ``accurately'' the interpretation represents the true reasoning process of the model.

A common pattern of achieving faithfulness in interpretation of neural models is via decomposition of a model into steps and inspecting the intermediate steps \cite[cognitive chunks;][]{doshi2017rigorous-humangrounded}. For example, neural modular networks \cite[NMN;][]{andreas16nmn-for-qa} first build an execution graph out of neural building blocks, and then apply this graph to data. The graph structure is taken to be a faithful interpretation of the model's behavior, as it describes the computation precisely. Similarly, \emph{highlight} methods (also called \emph{extractive rationales}\footnotemark), decompose a textual prediction problem into first \emph{selecting} highlighted texts, and then \emph{predicting} based on the selected words (\emph{select-predict}, described in Section \ref{sec:background}). The output of the selection component is taken to be a faithful interpretation, as we know exactly which words were selected. Similarly, we know that words that were not selected do not participate in the final prediction.

However, \citet{sanjay2020} call NMN graphs \emph{not faithful} in cases where there is a discrepancy between a building block's behavior and its name (i.e. expected behavior). Can we better characterize the requirement of faithful interpretations and amend this discrepancy?

\footnotetext{
The term `rationale' \cite{lei16} is more commonly used for this format of explanation in NLP, for historical reasons: highlights were associated with human rationalization of data annotation \citet{DBLP:conf/naacl/ZaidanEP07}. We argue against widespread use of this term, as it refers to multiple concepts in NLP and ML
 \cite[e.g.,][]{DBLP:conf/naacl/ZaidanEP07,DBLP:conf/emnlp/BaoCYB18,eraser2019,10.1145/3377325.3377512}, and importantly, `rationalization' attributes human intent to the highlight selection which is not necessarily compatible with the model, as we show in this work.
}

We take an extensive and critical look at the formalization of faithfulness and of explanations, with textual highlights as an example use-case. In particular, the \textit{select-predict} formulation for faithful highlights raises more questions than it provides answers: we describe a variety of curious failure cases of such models in Section~\ref{sec:limitations}, as well as experimentally validate that the failure cases are indeed possible and do occur in practice. Concretely, the behavior of the selector and predictor in these models do not necessarily line up with expectations of people viewing the highlight. Current literature in ML and NLP interpretability fails to provide a theoretical foundation to characterize these issues (Sections~\ref{sec:limitations}, \ref{sec:plausibility}).

To remedy this, we turn to literature on the science of social explanations and how they are utilized and perceived by humans (Section~\ref{sec:social}): the social and cognitive sciences find that human explanations are composed of two, equally important parts: the attribution of a causal chain to the decision process (\textit{causal attribution}), and the attribution of social or human-like intent to the causal chain (\textit{social attribution}) \cite{miller2017social}, where `human-like intent' refers to a system of beliefs and goals behind or following the causal process. E.g., ``she drank the water \emph{because she was thirsty}.''\footnote{Note that coincidence (lack of intent) is also a part of this system.} 

People may also \emph{attribute social intent to models}: in the context of NLP, when observing that a model consistently translates ``doctor'' with male morphological features \cite{stanovsky-etal-2019-evaluating}, the user may attribute the model with a ``belief'' all doctors are male, despite the model lacking an explicit system of beliefs. Explanations can influence this social attribution: for example, a highlight-based explanation may influence the user to attribute the model with the intent of ``performing a summary before making a decision'' or ``attempting to justify a prior decision''.

Fatally, the second key component of human explanations---the social attribution of intent---has been missing from current formalization on the desiderata of artificial intelligence explanations. 
In Section~\ref{sec:perceived-definition} we define that a faithful interpretation---a causal chain of decisions---is \textit{aligned} with human expectations if it is adequately constrained by the social behavior attributed to it by human observers.

Armed with this knowledge, we can now verbalize the issue behind the ``non-faithfulness'' perceived by \citet{sanjay2020} for NMNs: the inconsistency between component names and their actual behavior caused a misalignment between the causal and social attributions. We can also characterize the more subtle issue underlying the failures of the
\textit{select-predict} models described in Section~\ref{sec:limitations}: In Section~\ref{sec:highlight-attribution} we argue that for a set of possible social attributions, the \textit{select-predict} formulations fails to guarantee \emph{any} of them. 

In Section~\ref{sec:highlight-as-evidence} we propose an alternative causal chain for highlights explanations: \textit{predict-select-verify}.
Predict-select-verify does not suffer from the issue of misaligned social attribution, as the highlights can only be attributed as \textit{evidence} towards the predictor's decision.   As a result, predict-select-verify highlights do not suffer from the misalignment failures of Section~\ref{sec:limitations}, and guarantee that the explanation method does not reduce the score of the original model.

Finally, in Section~\ref{sec:contrastive} we discuss an implementation of \textit{predict-select-verify}, i.e., designing the components in the roles predictor and selector. Designing the selector is non-trivial, as there are many possible options to select highlights that evidence the predictor's decision, and we are only interested in selecting ones that are meaningful for the user to understand the decision. We leverage observations from cognitive research regarding the internal structure of (human-given) explanations, dictating that explanations must be \textit{contrastive} to hold tangible meaning to humans. We propose a classification \textit{predict-select-verify} model which provides contrastive highlights---to our knowledge, a first in NLP---and qualitatively exemplify and showcase the solution.

\paragraph{Contributions.} We identify shortcomings in the definitions of faithfulness and plausibility to characterize what is useful explanation, and argue that the \emph{social attribution} of an interpretation method must be taken into account.
We formalize ``aligned faithfulness'' as the degree to which the causal chain is aligned with the social attribution of intent that humans perceive from it. Based on the new formalization, we (1) identify issues with current \textit{select-predict} models that derive faithful highlight interpretations, and (2) propose a new causal chain that addresses these issues, termed \textit{predict-select-verify}. Finally, we implement this chain with \textit{contrastive explanations}, previously unexplored in NLP explanaibility. We make our code available online.\footnote{\url{https://github.com/alonjacovi/aligned-highlights}}


\section{Highlights as Faithful Interpretations} \label{sec:background}

Highlights, also known as extractive rationales,
are binary masks over a given input which imply some behavioral interpretation (as an incomplete description) of a particular model's decision process to arrive at a decision on the input. 


Given input sequence $x \in \mathds{R}^n$ and model $m:~\mathds{R}^n~\longrightarrow~Y$, a highlight interpretation $h \in \mathds{Z}_2^n$ is a binary mask over $x$ which attaches a meaning to $m(x)$, where the portion of $x$ highlighted by $h$ was important to the decision.




This functionality of $h$ was interpreted by \citet{lei16} as an implication of a behavioral process of $m(x)$, where the decision process is a modular composition of two unfolding stages:
\begin{enumerate}
    \item \textbf{Selector} component $m_s: \mathds{R}^n \longrightarrow \mathds{Z}_2^n$ selects a binary highlight $h$ over $x$.
    \item \textbf{Predictor} component $m_p: \mathds{R}^n \longrightarrow Y$ makes a prediction on the input $h \odot x$.
\end{enumerate}

The final prediction of the system at inference is $m(x) = m_p(m_s(x) \odot x)$. We refer to $h := m_s(x)$ as the \textbf{highlight} and $h \odot x$ as the \textbf{highlighted text}.

What does the term `faithfulness' mean in this context? A highlight interpretation can be \textit{faithful} or \textit{unfaithful} to a model. Literature accepts a highlight interpretation as `faithful' if the highlighted text was provably the only input to the predictor.


\paragraph{Implementations.}
Various methods have been proposed to train select-predict models. Of note: \citet{lei16} propose to train the selector and predictor end-to-end via  REINFORCE \cite{williams1992simple}, and  \citet{bastings-etal-2019-interpretable} replace REINFORCE with the reparameterization trick \cite{kingma-vae}. \citet{jain2020} propose FRESH, where the selector and predictor are trained separately and sequentially.

\section{Use-cases for Highlights}



To discuss whether an explanation procedure is useful as a description of the model's decision process, we must first discuss what is considered useful for the technology.

We refer to the following use-cases:
\\[0.15em]
\noindent\textbf{Dispute:} A user may want to dispute a model's decision, e.g. in a legal setting. They can do this by disputing the selector or predictor: by pointing to some non-selected words, saying: ``the model wrongly ignored \textit{A},'' or by pointing to selected words and saying: ``based on this highlighted text, I would have expected a different outcome.''
\\[0.3em]
\noindent\textbf{Debug:} Highlights allow the developer to designate model errors into one of two categories: did the model focus on the wrong part of the input, or did the model make the wrong prediction based on the correct part of the input? Each category implies a different method of alleviating the problem.
\\[0.3em]
\noindent\textbf{Advice}: Assuming that the user is unaware of the ``correct'' decision, they may (1) elect to trust the model, and learn from feedback on the part of the input relevant to the decision; or (2) elect to increase or decrease trust in the model, based on whether the highlight is aligned with the user's prior on what the highlight should or should not include. E.g., if the highlight is focused on punctuation and stop words, while the user believes it should focus on content words.

\section{Limitations of Select-Predict Highlights as (Faithful) Interpretations} \label{sec:limitations}



We now explore a variety of circumstances 
in which select-predict highlight interpretations are \textbf{uninformative} to the above use-cases.
While we present these failures in context of highlights, they should be understood generally as symptoms of mistaken or missing formal perspective of explanations in machine learning. Specifically, that \textbf{faithfulness is insufficient to formalize the desired properties of \add{interpretations}.}



\subsection{Trojan Explanations} \label{subsec:trojan}


Task information can manifest in the interpretation in exceedingly unintuitive ways, making it faithful, but functionally useless. 
We lead with an example, to be followed by a more exact definition.

\paragraph{Leading example.} Consider the following case of a faithfully highlighted decision process:

\begin{enumerate}[noitemsep]
    \item The selector makes a prediction $y$, and encodes $y$ in a highlight pattern $h$. It then returns the highlighted text $h\odot x$.
    \vspace{0.1em}
    \item The predictor recovers $h$ (the binary mask vector) from $h \odot x$ and decodes $y$ from the mask pattern $h$, \textit{without} relying on the text.
\end{enumerate}

The selector may choose to encode the predicted class by the location of the highlight (beginning vs. end of the text), or as text with more highlighted tokens than non-highlighted, etc.
This is problematic: while the model appears to make a decision based on the highlight's word content, the functionality of the highlight serves a different purpose entirely.

Evidently, this highlight ``explains'' nothing of value to the user. The role of the highlight is completely misaligned with the role expected by the user. E.g., in the \textit{advice} use-case, the highlight may appear random or incomprehensible. This will cause the user to lose trust in the prediction, even if the model was making a reasonable and informed decision. 

This case may seem unnatural and unlikely. Nevertheless, it is \textit{not explicitly avoided} by faithful highlights of \textit{select-predict} models: this ``unintentional'' exploit of the modular process is a valid trajectory in the training process of current methods. We verify this by attempting to predict the model's decision based on the mask $h$ alone via another model (Table~\ref{tab:trojan-results-h}). The experiment surprisingly succeeds at above-random chance. Although the result does not ``prove'' that the predictor uses this unintuitive signal, it shows that \textit{there is no guarantee that it doesn't}.

\begin{table}[t]
\centering
    \resizebox{0.98\linewidth}{!}{%
\begin{tabular}{lcccccc}
\toprule
              Model & SST-2 & AGNews & IMDB & Ev.Inf. & 20News & Elec \\
               \midrule
Random baseline & 50.0 & 25.0 & 50.0 & 33.33 & 5.0 & 50.0 \\
\citet{lei16}       & 59.7 & 41.4 & 69.11 & 33.45 & & 60.75  \\
\citet{bastings-etal-2019-interpretable}  & 62.8 & 42.4 & & & 9.45 &  \\
FRESH & 52.22 & 35.35 & 54.23 & 38.88 & 11.11 & 58.38 \\
\bottomrule
\end{tabular}}
\caption{The performance of an RNN classifier using $h$ alone as input, in comparison to the random baseline. Missing cells denote cases where we were unable to converge training.}
\label{tab:trojan-results-h}
\end{table}

\begin{table}[t]
\centering
    \resizebox{0.98\linewidth}{!}{%
\begin{tabular}{lccccc}
\toprule
              Model & AGNews & IMDB & 20News & Elec \\
               \midrule
Full text baseline & 41.39 & 51.22 & 8.91 & 56.41 \\ 
\citet{lei16}       & 46.83 & 57.83 & & 60.4  \\
\citet{bastings-etal-2019-interpretable}  & 47.69 & & 9.66 \\
FRESH & 43.29 & 52.46 & 12.53 & 57.7 \\
\bottomrule
\end{tabular}}
\caption{The performance of a classifier using quantities of the following tokens in $h \odot x$: comma, period, dash, escape, ampersand, brackets, and star; as well as the quantity of capital letters and $|h \odot x|$. Missing cells are cases where we were unable to converge training.}
\label{tab:trojan-results-statistics}
\end{table}

\paragraph{Definition (Trojan Explanations).} We term the more general phenomenon demonstrated by the example a \textit{Trojan explanation}: the explanation (in our case, $h \odot x$) carries information which is encoded in ways that are `unintuitive' to the user which observes the interpretation as an explanation of model behavior. \add{This means that the users observing the explanation naturally expects the explanation to convey information in a particular way\footnote{The expectation of a `particular way' is defined by the attribution of intent, explored later (\S\ref{sec:perceived-definition}). Precise descriptions of this expectation depend on the model, task, and user.} which is different from the true mechanism of the explanation.}


\begin{table*}[t]
\centering
    \resizebox{0.98\linewidth}{!}{%
\begin{tabular}{lp{20cm}l}
\toprule
              Model & Text and Highlight & Prediction \\
               \midrule
(a) & i really don't have much to say about this book holder, not that it's just a book holder. it's a \hl{nice} one. it does it's job . it's a little too expensive for just a piece of plastic. it's strong, sturdy, and it's big enough, even for those massive heavy textbooks, like the calculus ones. although, i would not recommend putting a dictionary or reference that's like 6'' thick (even though it still may hold). it's got little clamps at the bottom to prevent the page from flipping all over the place, although those tend to fall off when you move them. but that's no big deal. just put them back on. this book holder is kind of big, and i would not put it on a small desk in the middle of a classroom, but it's not too big. you should be able to put it almost anywhere when studying on your own time. & Positive \\ \midrule
(b) & i really don't have much to say about this book holder, not that it's just a book holder. \hl{it's a nice one. it does it's job .} it's a \hl{little too expensive} for just a piece of plastic. it's strong, sturdy, and it's big enough, even for those massive heavy textbooks, like the calculus ones. although, i would not recommend putting a dictionary or reference that's like 6'' thick (even though it still may hold). it's got little \hl{clamps} at the bottom to prevent the page from flipping all over the place, although those \hl{tend to fall off} when you move them. \hl{but that's no big deal.} just put them back on. this book holder \hl{is kind of big}, and i would not put it on a small desk in the middle of a classroom, \hl{but it's not too big.} you should be able to put it almost anywhere when studying on your own time. & Positive \\
\bottomrule
\end{tabular}}
\caption{Highlights faithfully attributed to two fictional select-predict models on an elaborate Amazon Reviews sentiment classification example. Although highlight (a) is easier to understand, it is also far less useful, as the selector clearly made hidden decisions.}
\label{fig:case2-ex}
\end{table*}




The `unintuitive' information encoded in $h \odot x$ is not limited to $h$ itself, and can be anything which is useful to predict $y$ and which the user will be unlikely to easily comprehend.
To illustrate, we summarize cases of Trojans (in highlight interpretations) which are reasonably general to multiple tasks:
\begin{enumerate}[noitemsep]
    \item \textbf{Highlight signal}: The label is encoded in the mask $h$ alone, requiring no information from the original text it is purported to focus on.
    \item \textbf{Arbitrary token mapping}: The label is encoded via some mapping from highlighted tokens to labels which is considered arbitrary to the user: e.g., commas for one class, and periods for another; the quantity of capital letters; distance between dashes, and so on.
    \item \textbf{The default class}: In a classification case, a class can be predicted by precluding the ability to predict all other classes and selecting it by default. As a result, the selector may decide that the absence of class features in itself defines one of the classes.
\end{enumerate}

In Table~\ref{tab:trojan-results-statistics} we experiment with type (2) above: we predict the decision (via MLP classifier) of a model from quantities of various characters, such as commas and dashes, in the highlighted texts generated by the models.\footnote{We count the following characters for a feature vector of length 10: comma, period, dash, escape, ampersand, both brackets, quantity of capital letters and length (by tokens) of the highlighted text.}
We compare to a baseline of predicting the decisions based on the same statistics from the full text. Surprisingly, all models show an \textit{increased} ability to predict their decisions on some level compared to the baseline. 

\paragraph{Conclusion.} Trojan explanations are not merely possible, but \textit{just as reasonable} to the model as any other option unless countered explicitly.
However, explicit modeling of Trojans is difficult, as our definition depends on user perception and contains limitless possible exploits. Even more critical, our current formal definition of what constitutes a faithful explanation does not rule our trojan explanations, and we cannot point to a property that makes such trojan explanations undesirable.

\subsection{The Dominant Selector} \label{subsec:dominant}

In another failure case, the selector makes an implicit decision, and proceeds to \textit{manipulate} the predictor towards this decision (without necessarily manifesting as a ``Trojan''). This means that \textbf{the selector can dictate the decision with a highlight that is detached from the selector's inner reasoning process.} 

While in the case of Trojan explanations the highlight's explanatory power is misunderstood by the user (but nevertheless exists), in this failure case, the information in the highlight is unproductive as an explanation altogether.

Suppose that the selector has made some decision based on some span A in the input, while producing span B to pass to the predictor---confident that the predictor will make the same prediction on span B as the selector did on span A. While span B may seem reasonable to human observers, it is a `malicious' manipulation of the predictor.

The dominant selector can realistically manifest when span A is more informative to a decision than span B, but the selector was incentivized, for some reason, to prefer producing span B over span A.
This is made possible because, while span B may not be a good predictor for the decision, it can become a good predictor \emph{conditioned on the existence of span A in the input.}
Therefore, \textbf{as far as the predictor is concerned, the probability of the label conditioned on span B is as high as the true probability of the label conditioned on span A.} We demonstrate with examples:

\paragraph{Example 1.} Consider the case of the \textit{binary sentiment analysis} task, where the model predicts the polarity of a particular snippet of text. Given this fictional movie review:

\vspace{0.5em}
\noindent ``Interesting movie about the history of \underline{\textit{Iran}}$^A$, only \underline{\textit{disappointed}}$^B$ that it's so short.''
\vspace{0.5em}

\noindent Assume that a select-predict model where the selector was trained to mimic human-provided rationales \cite{eraser2019}, and the predictor made a (mistaken) negative sentiment classification. Assume that \textit{Iran} (A) is highly correlated with negative sentiment, more-so than \textit{disappointed} (B)---as `not disappointed' and such are also common. Changing the word `Iran' to `Hawaii', for example, will change the prediction of the model from negative to positive. However, this correlation may appear controversial or unethical, and thus, humans tend to avoid rationalizing with it explicitly. The selector will be incentivized to make a negative prediction because of \textit{Iran} while passing \textit{disappointed} to the predictor. 

Since the choice of span B is conditioned on the choice of span A (meaning that the selector will choose \textit{disappointed} only if it had a-priori decided on the negative class thanks to \textit{Iran}), span B is just as informative to the predictor as span A is in predicting the negative label. 

This example is problematic not only due to the `interpretable' model behaving unethically, but \textbf{due to the inherent incentive of the model to lie and pretend it had made an innocent mistake} of overfitting to the word `disappointed'. 

\paragraph{Example 2.} Assume that two fictional select-predict models attempt to classify a complex, mixed-polarity review of a product. Table~\ref{fig:case2-ex} describes two fictional highlights faithfully attributed to these two models, on an example selected from the Amazon Reviews dataset.

The models make the same decision, yet their implied reasoning process is wildly different, thanks to the different highlight interpretations: model (a)'s selector made some decision and selected a word, ``nice'', which trivially supports that decision. The predictor, which can only observe this word, simply does as it is told. 
Comparatively, the selector of model (b) performed a very different job: as a summarizer. The predictor then made an informed decision based on this summary. How the predictor made its decision is unclear, but the \textit{division of roles} in (b) is significantly easier to comprehend---since the user expects the predictor to make the decision based on the highlight.

This has direct practical implications: in the \textit{dispute} use-case, given a dispute claim such as ``the sturdiness of the product was not important to the decision'', the claim appears impossible to verify in (a). The true decision may have been influenced by words which were not highlighted. The claim \emph{appears to be} safer in (b). But why is this the case? 

\begin{table*}[t]
\centering
    \resizebox{0.9\linewidth}{!}{%
\begin{tabular}{lccccccccc}
\toprule
               & SST-2 & SST-3   & SST-5           & AG News & IMDB  & Ev. Inf. & MultiRC & Movies & Beer \\
               \midrule
\citet{lei16}       & \textit{22.65} & \textit{7.09}    & 9.85                & 33.33   &  \textit{22.23}    &               36.59    & 31.43   & 160.0  &  37.93          \\
\citet{bastings-etal-2019-interpretable}  &   \textit{3.31}    &   \textit{ 0 }    & 2.97    &  199.02  & \textit{12.63}     &         85.19   & 75.0  &        &   13.64   \\
FRESH & 90.0  & \textit{17.82} & \textit{13.45}   & 50.0    & \textit{14.66}  & 9.76     & 0.0     & 20.0   &                         \\
\bottomrule
\end{tabular}}
\caption{The percentage \emph{increase} in error of selector-predictor highlight methods compared to an equivalent architecture model which was trained to classify complete text. We report the numbers reported in previous work whenever possible (\textit{italics} means our results). Architectures are \textit{not} necessarily consistent across the table, thus they do not imply performance superiority of any method.  The highlight lengths chosen for each experiment were chosen with precedence whenever possible, and otherwise chosen as 20\% following \citet{jain2020} precedence.}
\label{tab:compromise}
\end{table*}

\subsection{Loss of Performance} \label{subsec:loss-of-performance}

It is common for select-predict models to perform worse on a given task in comparison to models that classify the full text in ``one'' opaque step (Table \ref{tab:compromise}). Is this phenomenon a reasonable necessity of interpretability? 
Naturally, humans are able to provide highlights of decisions without any loss of accuracy.
Additionally, while interpretability may sometimes be prioritized over state-of-the-art performance, there are also cases that will disallow the implementation of artificial models unless they are strong \textit{and} interpretable.\footnote{For example, in the case of a doctor or patient seeking life-saving advice---it is difficult to quantify a trade-off between performance and explanation ability.}

We can say that there is some expectation for whether models \textit{can} or \textit{cannot} surrender performance in order to explain themselves. This expectation may manifest in one way or the other for a given interaction of explanation. And regardless of what this expectation may be in this scenario, select-predict models do follow the former (loss of performance exists) and human rationalization follows the latter (loss of performance does not exist), such that there is a clear mismatch between the two. How can we formalize if, or whether, this behavior of select-predict models is reasonable? What is the differentiating factor between the two situations?

\section{Explanatory Power of Highlights}

We have established three failure cases of select-predict highlights: Trojan explanations (\S\ref{subsec:trojan}) cause a misinterpretation of the highlight's functionality, and in dominant selectors (\S\ref{subsec:dominant}), the highlight does not convey any useful information. 
Finally, loss of performance (\S\ref{subsec:loss-of-performance}) shows an inherent, unexplained mismatch between the behavior of select-predict explainers and human explainers. 

All of these cases stem from a shared failure in formally characterizing the information to be conveyed to the user. 
For example, Trojan explanations are a symptom of the selector and predictor communicating through the highlight interface in an `unintended' manner; dominant selectors are a symptom of the selector making the highlight decision in an `unintended' manner, as well---but this is entirely due to the fact that we did not define what is intended, to begin with. Further, this is a general failure of interpretability, not restricted to select-predict highlight interpretations.


\section{On Faithfulness, Plausibility and Explanainability from the Science of Human Explanations} \label{sec:social}

The mathematical foundations of machine learning and natural language processing are insufficient to tackle the underlying issue behind the 
symptoms described in Section  \ref{sec:limitations}. In fact, formalizing the problem itself is difficult. What enables a faithful explanation to be ``understood'' as accurate to the model? And what causes an explanation to be perceived as a Trojan? 

In this section, we attempt to better formalize this problem on a vast foundation of social, psychological and cognitive literature about human explanations.\footnote{Refer to \citet{miller2017social} for a substantial survey in this area, which was especially motivating to us.} 


\subsection{Plausibility is not the Answer} \label{sec:plausibility}

\emph{Plausibility}\footnote{Refer to the supplementary material for a glossary of relevant terminology from the human explanation sciences.} (or \emph{persuasiveness}) is the property of an interpretation being convincing towards the model prediction, regardless of whether the model was correct or whether the interpretation is faithful. It is inspired by human-provided explanations as post-hoc stories generated to plausibly justify our actions \cite{rudin2019stop}. Plausibility is often quantified by the degree that the model's highlights resemble gold-annotated highlights given by humans \cite{bastings-etal-2019-interpretable,shiyuchang2020invariant-rationalization} or by querying for the feedback of people directly \cite{jain2020}.

Following the failure cases in Section~\ref{sec:limitations}, one may conclude that plausibility is a desirable, or even necessary, condition for a good interpretations: after all, Trojan explanations are by default implausible. We strongly argue this is not the case.

\textbf{Plausibility should be viewed an incentive of the explainer, and not as a property of the explanation:} human explanations can be categorized by utility across multiple axes
\cite{miller2017social}, among them are (1) \textit{learning} a better internal model for future decisions and calculations  \cite{lombrozo2006structure,williams2013hazards}; (2) \textit{examination} to verify the explainer has a correct internal prediction model; (3) \textit{teaching}\footnote{Although (1) and (3) are considered one-and-the-same in the social sciences, we disentangle them as that is only the case when the explainer and explainee are both human.} to modify the internal model of the explainer towards a more correct one (can be seen as the opposite end of (1)); (4) \textit{assignment of blame} to a component of the internal model; and finally, (5) \textit{justification} and \textit{persuasion}. These goals can be trivially mapped to our case, where the explainer is artificial. 

Critically, goal (5) of justification and persuasion by the explainer may not necessarily be the goal of the explainee. Indeed, in the case of AI explainability, it is \textit{not} a goal of the explainee to be persuaded that the decision is correct (even when it is), but to understand the decision process. If plausibility is a goal of the artificial model, this perspective outlines a game theoretic mismatch of incentives between the two players. And specifically in cases where the model is incorrect, it is interpreted as the difference between an innocent mistake and an intentional lie---of course, lying is considered more unethical. As a result, we conclude that \textbf{modeling and pursuing plausibility in AI explanations is an ethical issue}.

The failures discussed above do not stem from how (un)convincing the interpretation is, but from \textit{how well the user understands the reasoning process of the model}. If the user is able to comprehend the \emph{steps} that the model has taken towards its decision, then the user will be the one to decide whether these steps are plausible or not, based on how closely they fit the user's prior knowledge on whatever correct steps should be taken---regardless of whether the user knows the correct answer or whether the model is correct.

\subsection{The Composition of Explanations}

\citet{miller2017social} describes human explanations of behavior as a social interaction of knowledge transfer between the explainer and the explainee, and thus \textit{they are contextual}, and can be perceived differently depending on this context. Two central pillars of the explanation are \textit{causal attribution}---the attribution of a causal chain\footnote{See \citet{hilton05} for a breakdown of types of causal chains; we focus on unfolding chains in this work, but others may be relevant as well.} of events to the behavior---and \textit{social attribution}---the attribution of intent to others \cite{heider58}.

\textbf{Causal attribution describes faithfulness:} 
we note a stark parallel between causal attribution and faithfulness: for example, the select-predict composition of modules defines an unfolding causal chain where the selector hides portions of the input, causing the predictor to make a decision based on the remaining portions. In fact, recent work has connected an accurate (faithful) attribution of causality with increased explainability \cite{feder2020causalm,madumal2020causalreinforcement}.

Additionally, \textbf{social attribution is missing:}
\citet{heider44} describe an experiment where participants attribute human concepts of emotion, intentionality and behavior to animated shapes. Clearly, the same phenomenon persists when humans attempt to understand the predictions of artificial models: we naturally attribute social intent to artificial decisions.

Can models be constrained to adhere to this attribution?
Although informally, prior work on highlights has considered such factors before. \citet{lei16} describe desiderata for highlights as being ``short and consecutive'', and \citet{jain2020} interpreted ``short'' as ``around the same length as that of human-annotated highlights''. We assert that the nature of these claims is an attempt to constrain highlights to the social behavior implicitly attributed to them by human observers in the select-predict paradigm (discussed later).


\section{(Socially) Aligned Faithfulness}  \label{sec:perceived-definition}

Unlike human explainers, artificial explainers can exhibit a misalignment between
the causal chain behind a decision and the social attribution attributed to it. This is because the artificial decision process may not resemble human behavior.


By presenting to the user the causal pipeline of decisions in the model's decision process as an interpretation of this process, the user naturally conjures social intent behind this pipeline. 
In order to be considered comprehensible to the user, the attributed social intent must match the actual behavior of the model. This can be formalized as a set of constraints on the possible range of decisions at each step of the causal chain.

We claim that this problem is the root cause behind the symptoms in Section~\ref{sec:limitations}. Here we define the general problem independently from the narrative of highlight interpretations.

\paragraph{Definition.}
We say that an interpretation method is \emph{faithful} if it accurately describes causal information about the decision process of the decision. We say that the faithful method is \emph{human-aligned} (short for ``aligned with human expectations of social intent'') if the model and method adhere to the social attribution of intent by human observers. 

\add{Conversely, `\textit{misaligned}' interpretations are interpretations whose mechanism of conveying causal information is different from the mechanism that the user utilizes to glean causal information from the interpretation, where the mechanism is defined by the user's social attribution of intent towards the model.} Furthermore, we claim that this attribution (expected intent) is heavily dependent on the model's task and use-case: different use cases may call for different alignments.

\section{Alignment of \textit{Select-Predict} Highlight Interpretations}

\label{sec:highlight-attribution}


In contrast to the human-provided explanations, in the ML setup, our situation is unique in that
we have control over the causal chain but not the social attribution. Therefore, the social attribution must \emph{lead} the design of the causal chain.
In other words, we argue that we must first identify the behavior expected of the decision process, and then constrain the process around it.

\subsection{Attribution of Highlight Interpretations} \label{sec:social-attribution-of-highlights-from-psychology}

\begin{table}[]
    \centering
    \resizebox{0.98\linewidth}{!}{%
    \begin{tabular}{c p{8.7cm}}
        \toprule
         \# & Claim \\
         \midrule
         1 & The marked text is supportive of the decision. \\
        2 & The marked text is selected after making the decision. \\
    3 &  The marked text is sufficient for making the decision. \\
    4 & The marked text is selected irrespective of the decision. \\
    5 & The marked text is selected prior to making the decision. \\
    6 & The marked text includes all the information that informed the decision. \\
    \midrule 
    7 & The marking pattern alone is sufficient for making the decision by the predictor. \\
    8 & The marked text provides no explanation whatsoever. \\
    \bottomrule
    \end{tabular}}
    \caption{A list of claims which attribute causality or social intent to the highlight selection.}
    \label{tab:claims}
\end{table}

In order to understand what ``went wrong'' in the failure cases above, we need to understand what are \textit{possible} expectations---potential instances of social attribution---to the `rationalizing' select-predict models. Table~\ref{tab:claims} outlines possible claims that could be attributed to a highlight explanation. Claims 1-6 are claims which could reasonably be attributed to highlights, while claims 7-8 are \textit{not} likely to manifest.



These claims can be packaged as two high-level ``behaviors'':\\
\textbf{\textit{Summarizing} (3-6)}, where the highlight serves
as an extractive summary of the most important and useful parts of the complete text. The highlight is merely considered a compression of the text, with sufficient information to make informed decisions in a different context, towards some concrete goal. It is not selected with an answer in mind, but in anticipation that an answer will be derived in the future, for a question that has not been asked yet.
\\And \textbf{\textit{evidencing} (1-3)}, in which the highlight serves as supporting evidence towards a \textit{prior} decision that was not necessarily restricted to the highlighted text. 





Claims 7-8 are representative of the examples of failure cases discussed in Section~\ref{sec:limitations}.

\subsection{Issues with \textit{Select-Predict}} \label{sec:select-predict-misalignment}

We argue that select-predict is inherently misleading.
\textbf{While claims 1-6 are plausible attributions to select-predict highlights, \emph{none of them can be guaranteed} by a select-predict system}, in particular for systems in which the selector and predictor are exposed to the end task during training. 

If the select-predict system acts ``as intended'', selection happens before prediction, which is incompatible with claims 1-2. However, as we do not have control over the selector component, it cannot be guaranteed that the selector will not perform an implicit decision prior to its selection, and once a selector makes an implicit decision, the selected text becomes disconnected from the explanation. 
For example, the selector decided on class A, and chose span B because it ``knows'' this span will cause the predictor to predict class A (see Section \ref{subsec:dominant}).

In other words, the advertised select-predict chain may implicitly become a `predict-select-predict' chain. The first and hidden prediction step makes the final prediction step disconnected from the cause of the model's decision, because the second prediction is conditioned on the first one. 
This invalidates attributions 3-6. It also allows for 7-8.

Select-predict models are closest to the characteristics of highlights selected as summaries by humans---therefore they can theoretically be aligned with summary attribution if the selector is designed as a truly summarizing component, and has no access to the end-task. This is hard to achieve, and no current model has this property. 

The issues of Section~\ref{sec:limitations} are direct results of the above conflation of interests: 
Trojan highlights and dominant selectors result from a selector that makes hidden and unintended
decisions, so they serve as neither summary nor evidence towards the predictor's decision. Loss of performance is due to the selector acting as an imperfect summarizer---whether summary is relevant to the task to begin with, or not (as is the case in agreement classification, or natural language inference).

\section{\textit{Predict-Select-Verify}}
 \label{sec:highlight-as-evidence}


We propose the \textit{predict-select-verify} causal chain as a solution that can be constrained to provide highlights as evidence \add{(i.e., guarantee claims 1-3). This framework solves the misalignment problem by allowing the derivation of faithful highlights aligned with evidencing social attribution. }

The decision pipeline is as follows:

\begin{enumerate}
    \setlength{\itemsep}{0pt}
    \setlength{\parskip}{0pt}
    \item The predictor $m_p$ makes a prediction $\hat{y}:=m_p(x)$ on the full text. \vspace{0.3em}
    \item The selector $m_s$ selects $h := m_s(x)$ such that $m_p(h \odot x) = \hat{y}$.
\end{enumerate}

In this framework, the selector provides evidence which is verified to be useful to the predictor towards a particular decision. Importantly, the final decision has been made on the full text, and the selector is constrained to provide a highlight that adheres to this exact decision. The selector does not purport to provide a highlight which is comprehensive of all evidence considered by the predictor, but it provides a \textit{guarantee} that the highlighted text is supportive of the decision.

\paragraph{Causal attribution.} The selector highlights are provably faithful to the predict-select-verify chain of actions. They can be said to be \textit{faithful by construction} \cite{jain2020}, similarly to how select-predict highlights are considered faithful---the models undergo the precise chain of actions that is attributed to their highlights.

\paragraph{Social attribution.} The term ``rationalization'' fits the current causal chain, unlike in select-predict, and so there is no misalignment: the derived highlights adhere to the properties of highlights as evidence described in Section~\ref{sec:social-attribution-of-highlights-from-psychology}. The highlight selection is made under constraints that the highlight serve the predictor's prior decision, which is \textit{not} caused by the highlighted text. The constraints are then verified at the \textit{verify} step.

\paragraph{Solving the failure cases (\S\ref{sec:limitations}).} As a natural but important by-product result of the above, predict-select-verify addresses the failures of Section~\ref{sec:limitations}: Trojan highlights and dominant selectors are impossible, as the selector is constrained to only provide ``retroactive'' selections towards a specific priory-decided prediction. The selector cannot cause the decision, since it was made without its intervention. Finally, the highlights inherently cannot cause loss of performance, since they merely support a decision which was made based on the full text.


\section{Constructing a \textit{Predict-Select-Verify} Model with Contrastive Explanations} \label{sec:contrastive}

In order to design a model adhering to the predict-select-verify chain, we require solutions for the predictor and for the selector.

The \textbf{predictor} is constrained to be able to accept both full-text inputs and highlighted inputs. For this reason, we use masked language modeling (MLM) models, such as BERT \cite{devlin2018bert}, fine-tuned on the downstream task. The MLM pre-training is performed by recovering partially masked text, which conveniently suits our needs. We additionally provide randomly-highlighted inputs to the model during fine-tuning.

The \textbf{selector} is constrained to select highlights for which the predictor made the same decision as it did on the full text. However, there are likely many possible choices that the selector may make under this constraints, as there are many possible highlights that all result in the same decision by the predictor. \textbf{We wish for the selector to select \textit{meaningful} evidence to the predictor's decision.}\footnotemark \ What is meaningful evidence? To answer this question, we again refer to cognitive science on necessary attributes of explanations that are easy to comprehend by humans. \add{We stress that selecting meaningful evidence is critical for predict-select-verify to be useful.}

\footnotetext{For example, the word ``nice'' in Table \ref{fig:case2-ex}a is likely not useful supporting evidence, since it is a rather trivial claim, even in a predict-select-verify setup.}


\subsection{Contrastive Explanations}

\begin{table*}[t]
\centering
    \resizebox{0.98\linewidth}{!}{%
\begin{tabular}{lp{13cm}cccc}
\toprule
              \multirow{2}{*}{Procedure} & \multirow{2}{*}{Text and Highlight} & Label & Prediction & Foil Prediction & Contrast Prediction \\
               &  & $y$  & $m_p(x)$ & $m_p(h \odot x)$ &  $m_p(h_c \odot x)$ \\
              
               \midrule
Manual & Ohio Sues Best Buy, Alleging Used Sales (AP): AP - Ohio authorities \textbf{\hl{sued}} Best Buy Co. Inc. on Thursday, alleging the electronics retailer engaged in unfair and \hl{deceptive business practices}. & Business & Sci/Tech & Business & Sci/Tech \\  \midrule

Manual & HK Disneyland Theme Park to Open in September: Hong Kong's Disneyland theme park will open on Sept. 12, 2005 and become the driving force for \hl{growth} in the city's \hl{tourism industry}, Hong Kong's \textcolor{black}{\hl{\textbf{government}}} and Walt Disney Co. & Business & World & Business & World \\  \midrule

Manual & Poor? Who's poor? Poverty is down: The proportion of people living on less than \textcolor{black}{\hl{\textbf{\$}}}1 a day decreased from 40 to 21 percent of the \hl{global population} between 1981 and 2001, says the World Bank's latest annual report. & World & Business & World & Business \\  \midrule

Manual & \textcolor{black}{\hl{\textbf{Poor}}}? Who's poor? Poverty is down: The proportion of people living on less than \$1 a day decreased from 40 to 21 percent of the \hl{global population} between 1981 and 2001, says the World Bank's latest annual report. & World & Business & World & Business \\  \midrule

Manual & Poor? Who's poor? Poverty is down: The proportion of people living on less than \$1 a day \textcolor{black}{\hl{\textbf{decreased}}} from 40 to 21 percent of the \hl{global population} between 1981 and 2001, says the World Bank's latest annual report. & World & Business & World & Business \\ \midrule

Automatic & \textcolor{black}{\hl{\textbf{Siemens}}} Says Cellphone Flaw May Hurt Users and Its Profit: Siemens\hl{, the world's fourth-largest maker of mobile phones, said Friday that a software flaw that can create a piercing ring in its newest phone models might hurt earnings in its handset division.} & Business & Sci/Tech & Business & Sci/Tech \\ \midrule

Automatic & Siemens Says \textcolor{black}{\hl{\textbf{Cell}}}phone Flaw May Hurt Users and Its Profit: Siemens\hl{, the world's fourth-largest maker of mobile phones, said Friday that a software flaw that can create a piercing ring in its newest phone models might hurt earnings in its handset division.} & Business & Sci/Tech & Business & Sci/Tech \\ \midrule

Automatic & Siemens Says Cellphone Flaw May Hurt \textcolor{black}{\hl{\textbf{Users}}} and Its Profit: Siemens\hl{, the world's fourth-largest maker of mobile phones, said Friday that a software flaw that can create a piercing ring in its newest phone models might hurt earnings in its handset division.} & Business & Sci/Tech & Business & Sci/Tech \\ \midrule

Automatic & Siemens Says Cellphone Flaw May Hurt Users and Its Profit: \textcolor{black}{\hl{\textbf{Siemens}}}\hl{, the world's fourth-largest maker of mobile phones, said Friday that a software flaw that can create a piercing ring in its newest phone models might hurt earnings in its handset division.} & Business & Sci/Tech & Business & Sci/Tech \\

\bottomrule
\end{tabular}}
\caption{Examples of contrastive highlights (\S\ref{sec:contrastive}) of instances from the AG News corpus. The model used for $m_p$ is fine-tuned bert-base-cased. The foil highlight $h$ is in \hl{standard yellow}; the contrastive delta $h_\Delta$ is in \hl{\textbf{bold yellow}}; and $h_c := h + h_\Delta$. All examples are cases of model errors, and the foil was chosen as the gold label.
}
\label{fig:topic-ex}
\end{table*}

An especially relevant observation in the social science literature is of \textit{contrastive explanations} \cite{miller2017social,miller2020contrastive}, following the notion that the question ``why $P$?'' is followed by an addendum: ``why $P$, rather than $Q$?'' \cite{hilton1988logic}. We refer to $P$ as the \textit{fact} and $Q$ as the \textit{foil} \cite{lipton1990contrastive}. The concrete valuation in the community is that in the vast majority of cases, the cognitive burden of a ``complete'' explanation, i.e. where $Q$ is $\overline{P}$, is too great, and thus $Q$ is selected as a subset of all possible foils \cite{hilton1986knowledge,Hesslow1988}, and often not explicitly, but implicitly derived from context. 

For example, ``Elmo drank the water because he was thirsty,'' explains the fact ``Elmo drank water'' without mentioning a foil. But while this explanation is acceptable if the foil is ``not drinking'', it is not acceptable if the foil is ``drinking tea'': ``Elmo drank the water \textit{(rather than the tea)} because he was thirsty.'' Similarly, ``Elmo drank the water because he hates tea'' only answers the latter foil. The foil is implicit in both cases, but nevertheless it is not $\overline{P}$, but only a subset.

In classification tasks, the implication is that an interpretation of a prediction of a specific class is hard to understand, and should be contextualized by \textit{the preference of the class over another}---and the selection of the foil (the non-predicted class) is non-trivial, and a subject of ongoing discussion even in human explanations literature.

Contrastive explanations have many implications for explanations in AI as a vehicle for explanations that are easy to understand. Although there is a modest body of work on contrastive explanations in machine learning \cite{Dhurandhar2018ExplanationsBO,Chakraborti2019ContrastiveAF,Chen2020TowardsTR}, to our knowledge, the NLP community seldom discusses this format.

\subsection{Contrastive Highlights}


An explanation in a classification setting should not only addresses the fact (predicted class), but do so against a foil (some other class).\footnote{Selecting the foil, or selecting what to explain, is a difficult and interesting problem even in philosophical literature \cite{Hesslow1988,mcgill93,chin2017contrastive}. In the classification setting, it is relatively simple, as we may request the foil (class) from the user, or provide separate contrastive explanations for each foil.} Given classes $c$ and $\hat{c}$, where $m_p(x) = c$, we will derive a contrast explanation towards the question: ``why did you choose $c$, rather than $\hat{c}$?''.

We assume a scenario where, having observed $c$, the user is aware of some highlight $h$ which should serve, they believe, as evidence for class $\hat{c}$.
In other words, we assume the user believes a pipeline where $m_p(x) = \hat{c}$ and $m_s(x) = h$ is reasonable.

If $m_p(h \odot x) \neq \hat{c}$, then the user is made aware that the predictor disagrees that $h$ serves as evidence for $\hat{c}$.

Otherwise, $m_p(h \odot x) = \hat{c}$. We define:
  \begin{equation*}
     h_c := \argmin_{%
      \substack{%
         h + h_{\Delta} \\
         \text{s.\,t.}\, \, \, |h_{\Delta}| > 0  \\
         \land \, m_p((h + h_{\Delta}) \odot x) = c
      }
     }
     |h + h_{\Delta}|.
  \end{equation*}
  
$h_c$ is the minimal highlight containing $h$ such that $m_p(h_c \odot x) = c$.
Intuitively, the claim by the model is as such: ``I consider $h_{\Delta}$ as a sufficient change from $h$ (evidence to $\hat{c}$) to $h_c$ so that it is evidence towards $c$.''



The final manual procedure is, given a model $m_p$ and input $x$:
\begin{enumerate}[noitemsep]
    \item The user observes $m_p(x)$ and chooses a relevant foil $\hat{c} \neq m_p(x)$.
    \item The user chooses a highlight $h$ which they believe supports $\hat{c}$.
    \item If $m_p(h \odot x) \neq \hat{c}$, the shortest $h_c$ is derived such that $h \subset h_c$ and $m_p(h_c \odot x) = m(x)$ by brute-force search.
\end{enumerate}

\paragraph{Automation.} While we believe the above procedure is most useful and informative, we acknowledge the need for automation of it to ease the explanation process. Steps 1 and 2 involve human input which can be automated: in place of step 1, we may simply repeat the procedure separately for each of all foils (and if there are too many to display, select them with some priority and ability to switch between them after-the-fact); and in place of step 2, we may heuristically derive candidates for $h$---e.g. the longest highlight for which the model predicts the foil:
  \begin{equation*}
     h := \argmax_{%
      \substack{%
         h \\
         m_p(h \odot x) = \hat{c}
      }
     }
     |h|.
  \end{equation*}
The automatic procedure is then, for each class $\hat{c} \neq m_p(x)$:
\begin{enumerate}
\itemsep0em 
    \item Candidates for $h$ are derived, e.g., the longest highlight $h$ for which $m_p(h \odot x) = \hat{c}$.
    \item The shortest $h_c$ is derived such that $h \subset h_c$ and $m_p(h_c \odot x) = m_p(x)$.
\end{enumerate}

We show examples of both procedures 
in Table \ref{fig:topic-ex} on examples from the AG News dataset. For illustration purposes, we selected incorrectly-classified examples, and selected the foil to be the true label of the example. The highlight for the foil was chosen by us in the manual examples.

In the automatic example, the model made an incorrect \textit{Sci/Tech} prediction on a \textit{Business} example. The procedure reveals that the model would have made the correct prediction if the body of the news article was given without its title, and that the words ``Siemens'', ``Cell'' and ``Users'' in the title are independently sufficient to flip the prediction on the highlight from \textit{Business} to \textit{Sci/Tech}.

We stress that while the examples presented in these figures appear reasonable, the true goal of this method is not to provide highlights that seem justified, but to provide a framework which allows models to be meaningfully incorporated in use-cases of \textit{dispute}, \textit{debug}, and \textit{advice}, with robust and proven guarantees of behavior. 



For example, in each of the example use-cases:
\\[0.15em]
\noindent\textbf{Dispute:} The user verifies if the model ``correctly'' considered a specific portion of the input in the decision: the model made decision $c$, where the user believes decision $\hat{c}$ is appropriate and is supported by evidence $h \odot x$. If $m_p(h \odot x) \neq c$, they may dispute the claim that the model interpreted $h \odot x$ with ``correct'' evidence intent. Otherwise the dispute cannot be made, as the model provably considered $h$ as evidence for $\hat{c}$, yet insufficiently so when combined with $h_{\Delta}$ as $h_c \odot x$.
\\[0.3em]
\noindent\textbf{Debug:} Assuming $c$ is incorrect, the user performs error analysis by observing which part of the input is sufficient to steer the predictor away from the correct decision $\hat{c}$. This is provided by $h_{\Delta}$.
\\[0.3em]
\noindent\textbf{Advice}: When the user is unaware of the answer, and is seeking perspective from a trustworthy model: they are given explicit feedback on which part of the input the model ``believes'' is sufficient to overturn the signal in $h$ towards $\hat{c}$. If the model is not considered trustworthy, the user may gain or reduce trust by observing whether $m(h \odot x)$ and $h_{\Delta}$ align with user priors.


\section{Discussion}

\paragraph{Causal attribution of heat-maps.} Recent work on the faithfulness of attention heat-maps \cite{Baan2019DoTA,Pruthi2019LearningTD,sofia-isattentioninterpretable} or saliency distributions \cite{alvarez2018robustness,kindermans2017unreliability-of-saliency} cast doubt on their faithfulness as indicators to the significance of parts of the input (to a model decision). Similar arguments can be made regarding any explanation in the format of heat-maps, such as LIME and SHAP \cite{jacovi2020}. We argue that this is a natural conclusion from the fact that, as a community, we have not envisioned an appropriate causal chain that utilizes heat-maps in the decision process, reinforcing the claims in this work on the parallel between causal attribution and faithfulness. This point is also discussed at length by \citet{grimsley-etal-2020-attention}.

\paragraph{Social attribution of heat-maps.} As mentioned above, the lack of a clear perception of a causal chain behind heat-map feature attribution explanations in NLP makes it difficult to discuss the social intent attributed by these methods. Nevertheless, it is possible to do so under two perspectives: (1) when the heat-map is discretized into a highlight, and thus can be analyzed along the list of possible attributions in Table \ref{tab:claims}; or (2) when the heat-map is regarded as a collection of pair-wise claims about which part the input is more important, given two possibilities. Perspective (1) can be likened to claims \#1 and \#2 in Table \ref{tab:claims}, i.e., ``evidencing'' attributions sans sufficiency.


\paragraph{Contrastive explanations in NLP.} We are not aware of prior work that discusses or implements contrastive explanations explicitly in NLP, however this does not imply that existing explanation methods in NLP are not contrastive. To the contrary, the social sciences argue that \textit{every} manner of explanation has a foil, and is comprehended by the explainee against some foil---including popular methods such as LIME \cite{Ribeiro:2016:WIT:2939672.2939778} and SHAP \cite{SHAP_NIPS2017_7062}. The question then becomes \textit{what} the foil is, and \textit{whether this foil is intuitive} and thus useful. In the case of LIME, for example, the foil is defined by the aggregation of all possible perturbations admitted to the approximating linear model---where such perturbations may not be natural language, and thus less intuitive as foil; additionally, \citet{kumar2020shapley-problems} have recently derived the foil behind general shapley value-based explanations, and have shown that this foil is not entirely aligned with human intuition. \emph{We argue that making the foil explicit and intuitive is an important goal of any interpretation system.}

\paragraph{Inter-disciplinary research.} Research on explanations in artificial intelligence will benefit from a deeper inter-disciplinary perspective on two axes: (1) literature on causality and causal attribution, regarding causal effects in a model's reasoning process; and (2) literature on the social perception and attribution of human-like intent to causal chains of model decisions or behavior. 

\section{Related Work}


How interpretations are comprehended by people is related to \textit{simulatability} \cite{kim2017interpretability}---the degree to which humans can simulate model decisions. 
Quantifying simulatability \cite{hase2020simulatability} is decidedly different from social alignment, since e.g., it will not necessarily detect dominant selectors. We theorize that aligned faithful interpretations will increase simulatability.

\textit{Predict-select-verify} is reminiscent of iterative erasure \cite{feng2018pathologies}. By iteratively removing ``significant'' tokens in the input, \citeauthor{feng2018pathologies} show that a surprisingly small portion of the input could be interpreted as evidence for the model to make the prediction, leading to conclusions on the pathological nature of neural models and their sensitivity to badly-structured text. This experiment retroactively serves as a successful application of \textit{debugging} using our formulation.


The approach by \citet{DBLP:conf/nips/ChangZYJ19} for class-wise highlights is reminiscent of contrastive highlights, but nevertheless distinct, since such highlights still explain a fact against all foils.

\section{Conclusion}

Highlights are a popular format for explanations of decisions on textual inputs, for which there are models available today with the ability to derive highlights ``faithfully''. We analyze highlights as a case-study in pursuit of rigorous formalization of quality artificial intelligence explanations.

We redefine \emph{faithfulness} as the accurate representation of the causal chain of decision making in the model, and \emph{aligned faithfulness} as a faithful interpretation which is also aligned to the social attribution of intent behind the causal chain. 
The two steps of causal attribution and social attribution \textit{together} complete the process of ``explaining'' the decision process of the model to humans. 

With this formalization, we characterize various failures in faithful highlights that ``seem'' strange, but could not be properly described previously, noting they are not properly constrained by their social attribution as summaries or evidence. We propose an alternative which can be constrained to serve as evidence. 
Finally, we implement our alternative by formalizing \textit{contrastive explanations} in the highlight format.

\section*{Acknowledgements}

This project has received funding from the European Research Council (ERC) under the European Union's Horizon 2020 research and innovation programme, grant agreement No. 802774 (iEXTRACT).

\bibliography{tacl2018}
\bibliographystyle{acl_natbib}

\newpage

\appendix

\section{Glossary} \label{appendix:definitions}

This work is concerned with formalization and theory of artificial models' explanations. We provide a (non-alphabetical) summary of terminology and their definitions as we utilize them. We stress that these definitions are \textit{not} universal, as the human explanation sciences describe multiple distinct perspectives, and explanations in AI are still a new field.
\\[0.3em] 
\noindent
\textbf{Unfolding causal chain:} 
A path of causes between a set of events, in which a cause from event C to event E indicates that C must occur before E. 
\\[0.3em] 
\noindent
\textbf{Human intent:} An objective behind an action. In our context, reasoning steps in the causal chain are actions that can be attributed with intent.
\\[0.3em] 
\noindent
\textbf{Interpretation:} A (possibly lossy) mapping from the full reasoning process of the model to a human-readable format, involving some implication of a causal chain of events in the reasoning process.
\\[0.3em] 
\noindent
\textbf{Faithful interpretation:} An interpretation is said to be faithful if the causal chain it describes is accurate to the model's full reasoning process. 
\\[0.3em] 
\noindent
\textbf{Explanation:} A process of conveying causal information about a model's decision to a person. We assume that the explainee always attributes intent to the actions of the explainer.
\\[0.3em] 
\noindent
\textbf{Plausibility:} Incentive of the explainer to provide justifying explanation that appears convincing.

\end{document}